\documentclass[conference]{IEEEtran}
\usepackage[backend=biber,style=numeric,sorting=none]{biblatex}
\ifCLASSINFOpdf
  \usepackage[pdftex]{graphicx}
  \graphicspath{{images/}} 
\else
\fi
\usepackage{csquotes}
\usepackage{hyperref}

\begin{document}

%
\title{Visual Attention: Deep Rare Features}

\author{\IEEEauthorblockN{Matei Mancas}
\IEEEauthorblockA{University of Mons\\
Information, Signal and\\
Artificial Intelligence Lab\\
matei.mancas@umons.ac.be}
\and
\IEEEauthorblockN{Phutphalla Kong}
\IEEEauthorblockA{Institute of Technology of Cambodia\\
Department of Information and\\
Communication Engineering\\
phutphalla@itc.edu.kh}
\and
\IEEEauthorblockN{Bernard Gosselin}
\IEEEauthorblockA{University of Mons\\
Information, Signal and\\
Artificial Intelligence Lab\\
bernard.gosselin@umons.ac.be}}


%

\maketitle

\begin{abstract}
Human visual system is modeled in engineering field providing feature-engineered methods which detect contrasted/surprising/unusual data into images. This data is \enquote{interesting} for humans and leads to numerous applications. Deep learning (DNNs) drastically improved the algorithms efficiency on the main benchmark datasets. However, DNN-based models are counter-intuitive: surprising or unusual data is by definition difficult to learn because of its low occurrence probability. In reality, DNNs models mainly learn top-down features such as faces, text, people, or animals which usually attract human attention, but they have low efficiency in extracting surprising or unusual data in the images. 

In this paper, we propose a model called DeepRare2019 (DR) which uses the power of DNNs feature extraction and the genericity of feature-engineered algorithms. DR 1) does not need any training, 2) it takes less than a second per image on CPU only and 3) our tests on three very different eye-tracking datasets show that DR is generic and is always in the top-3 models on all datasets and metrics while no other model exhibits such a regularity and genericity. DeepRare2019 code can be found at \url{https://github.com/numediart/VisualAttention-RareFamily}.
\end{abstract}


%
\IEEEpeerreviewmaketitle

\section{Introduction: deep learning trouble}
\label{sec:intro}
The human visual system handles a huge quantity of incoming visual information and it cannot carry out multiple complex tasks in the same time on the whole visual field. This bottleneck \cite{donald1958} implies that it has an exceptional ability of sampling the surrounding world and pay attention to objects of interest. In computer vision, visual attention is modeled through the so-called saliency maps. The modeling of visual attention has numerous applications such as object detection, image segmentation, image/video compression, robotics, image re-targeting, visual marketing and so on \cite{mancas2016}.

Since the early 2000, numerous models of visual attention based on image features were provided. In this paper, they will be referred as \enquote{classical models}. While they can be very different, most of them have the same main philosophy: search for contrasted, rare, abnormal or surprising features within a given context. Among those models one may find seminal work of \cite{itti2000} or \cite{rose1999}, but also more recent work based on information processing such as AIM \cite{aim}. Finally, some models became a reference for classical models such as GBVS \cite{gbvs}, RARE \cite{rare2012}, BMS \cite{bms2013} or AWS \cite{aws}. 

With the arrival of the deep learning wave, most researchers have focused on Deep Neural Networks saliency which will be referred as \enquote{DNN-based} in this paper. DNN-based models triggered a revolution in terms of results on the main benchmark datasets such as MIT benchmark \cite{mit-saliency-benchmark} where DNN-based saliency models definitely outperformed classical models. The DNN-based models have been already used in several applications such as image and video processing, medical signal processing, big data analysis, and saliency modeling as well \cite{sun15}, \cite{zhao15}, \cite{qin15}, \cite{han14}, \cite{sun16}. Some of the DNN-based models became new references such as SALICON \cite{salicon2015}, MLNet \cite{mlnet2016} or SAM-ResNet \cite{sam2018}.

However, recently DNN-based models have been criticized for some drawbacks. They underestimate the importance of bottom-up attention \cite{kum17} which  indicates that they were mostly trained to detect the attractive top-down objects rather than detect saliency itself. In \cite{kong19} the authors found that if saliency models very precisely detect top-down features, they neglect a lot of bottom-up information which is surprising and rare, thus by definition difficult to learn. This shows that saliency cannot be learnt but instead objects which are often attended by human gaze (such as faces, text, bodies, etc.) are learnt and by the way, they are enough to provide good results on the main benchmarks. Recently, \cite{Kotseruba2019} introduced two novel datasets, one based on psycho-physical patterns (P\textsuperscript{3}) and one based on natural odd-one-out (O\textsuperscript{3}) stimuli. They showed that while DNN-based models are good in MIT dataset on natural images, their results drastically drop on P\textsuperscript{3} and O\textsuperscript{3}. This shows that in addition to not take into account low-level features, DNN-based models are not generic enough to adapt to new images which are different enough from the training dataset. 

In parallel to DNN-based models, DeepFeat \cite{deepFeat} or SCAFI \cite{scafi} deal with models where pre-trained deep features are used. Those models will be called \enquote{deep-features models} in this paper. However, they are not yet comparable to DNN-based models for general images datasets such as the MIT benchmark. 

Based on the new datasets in \cite{Kotseruba2019}, we provide a new deep-feature saliency model called DeepRare2019 mixing deep features and the philosophy of an existing classical model \cite{rare2012}. Efficient on all the datasets, with no need for any training, efficient in terms of computation even on CPU and easily usable on any DNN architecture. 

In a section \ref{sec:DeepRare2019} the DeepRare2019 saliency model is described. In section \ref{sec:valid} this model is tested on the datasets proposed in \cite{Kotseruba2019}. We finally discuss and conclude on the pertinence of the come back of the feature engineering models. 

\section{DeepRare2019 model}
\label{sec:DeepRare2019}

Our contribution is in mixing the simplicity of the idea of rarity computation to find the most salient features with the advantages of deep features extraction. Indeed, rare features attract human attention as they are surprising compared to the other features within the image. The resulting model is called DeepRare2019 (\textbf{DR}). This combination has the advantage to be fast (less than 1 second per image on CPU with a VGG16 feature extractor) and easy to adapt to any default DNN architectures (VGG19, ResNet, etc.).

\subsection{Deep features extraction}

A convolutional network is a great tool for feature extraction. When trained on a general dataset such as ImageNET, the network will extract a complete set of features that one finds in images at several scales (from very low-level in the first layers to very high level in the last ones). We decide here to use a VGG16 architecture with its default training on ImageNET dataset as a feature extractor, but any other architecture could be used as well. Our implementation is based on Keras framework to extract the convolutional layers and feature maps within those layers. We do not use (1) the pooling layers (as they are redundant with the previous convolutional layer) and (2) the final fully connected classification layers. An example for layer 1 is illustrated in Figure \ref{fig:DR}.

In a VGG16, the convolutional layers are gathered within 5 groups separated by the pooling layers : 1) the first low-level features in layers 1 and 2, then 2) second set of low-level features from layers 4 and 5, after that 3) the first middle-level layers 7, 8 and 9 and 4) the second middle-level layers 11, 12 and 13 and finally 5) the high-level features from layers 15, 16 and 17.

\subsection{Rarity of deep features and top-down information}
On each feature map within the layers we compute the data rarity. For that we use the main idea from \cite{rare2012} without the multi-resolution part which is naturally achieved by the VGG16 architecture (and also by most of other architectures). A very simple rarity function \textit{R} based on the histogram of each feature map sampled on a few bins (11 in the current implementation) is used as in equation \ref{eqn1}.
\begin{eqnarray}
\label{eqn1}
   R(i) = -log(p(i))
\end{eqnarray}
where \textit{p(i)} is the occurrence probability for the pixels of bin \textit{i}. 
Once the rarity histogram \textit{R} is computed, the resulting rarity image is reconstructed by backprojection. This operation uses the histogram of a feature (here the rarity of a feature) and then use it to find this feature in an image projecting each histogram value on the corresponding pixel in an image. This image will highlight pixels in the feature map which are rare compared to the other pixels in the feature map. Based on \cite{rare2012}, rare pixels are the ones which might attract human attention. Rarity is applied on each feature map of each layer as it can be seen on the 64 feature maps of layer 1 in Figure \ref{fig:DR}.

The advantage of this approach is that it is very fast to compute and this is important as it needs to be applied to numerous feature maps.

\subsection{Data fusion}
\begin{figure}[!t]
\centering
\includegraphics[width=3.4in]{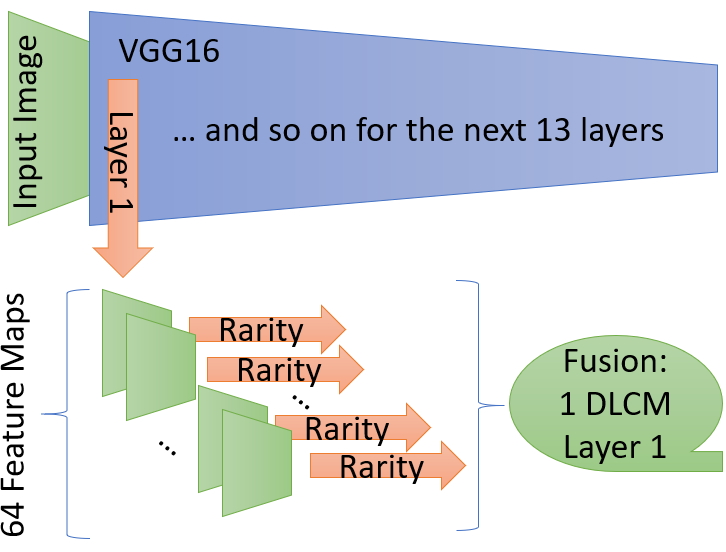}
\caption{Processing for Layer 1. This processing is iterated for all 13 convolutional layers from a VGG16 network.}
\label{fig:DR}
\end{figure}

Once the rarity of all feature maps is computed, the results need to be fused together. We use a classical map fusion from \cite{itti2004} where the fusion weights depend on the squared difference between the max and the mean of each map. This is applied to all feature maps within each layer leading to 13 deep layer conspicuity maps (DLCM), one for each convolutional layer in VGG16 (see Figure \ref{fig:DR} for first layer). 

In a second stage, the same fusion method is applied for each of the 5 layer groups arriving to 5 deep groups conspicuity maps (DGCM). This fusion is made in a way to give more importance to higher level layers. 

Finally, the 5 DGCM are summed up. A top-down face map is added based on feature map \#105 from layer 15 which is known to detect faces which are large enough \cite{scafi}.

\section{DeepRare2019 Validation}
\label{sec:valid}

\subsection{Data and Metrics for Validation}
\label{sec:db}
We use 3 datasets namely MIT1003 \cite{Judd2009}, P\textsuperscript{3}, and O\textsuperscript{3} datasets \cite{Kotseruba2019} to validate our results. The MIT dataset has general-purpose real-life images. P\textsuperscript{3} dataset evaluates the ability of saliency algorithms to find singleton targets which focuses on color, orientation, and size (without center bias). O\textsuperscript{3} dataset depicts a scene with multiple objects similar to each other in appearance (distractors) and a singleton (target) which focuses on color, shape, and size (with center bias). We decided to use these 3 very different datasets to check how saliency models behave when facing images in different contexts. 

Concerning metrics, we use measures from \cite{Kotseruba2019}. The \enquote{number of fixations} (\# fix.) is defined as the path formed by the saliency maximum followed by the other maxima of the saliency map before reaching the target. The global saliency index (GSI) measures how well the target mean saliency is distinguished from the distractors. The maximum saliency ratio (MSR) focuses on maximum saliency of the target versus the distractors \cite{Wloka2016} and the same for the background versus target (MSR\textsubscript{t} and MSR\textsubscript{b}). We also use standard eye-tracking evaluation metrics from MIT benchmark \cite{mit-saliency-benchmark} such as CC, KL, AUC Judd, AUC Borji, NSS, and SIM.

\subsection{Qualitative validation}
We compare our model to other models on P\textsuperscript{3} and O\textsuperscript{3} datasets. According to \cite{Kotseruba2019}, they observe that most classical models  perform better on P\textsuperscript{3} than DNN-based models. In contrast, DNN-based models perform better on O\textsuperscript{3}. 

\begin{figure}[!t]
\centering
\includegraphics[width=3.4in]{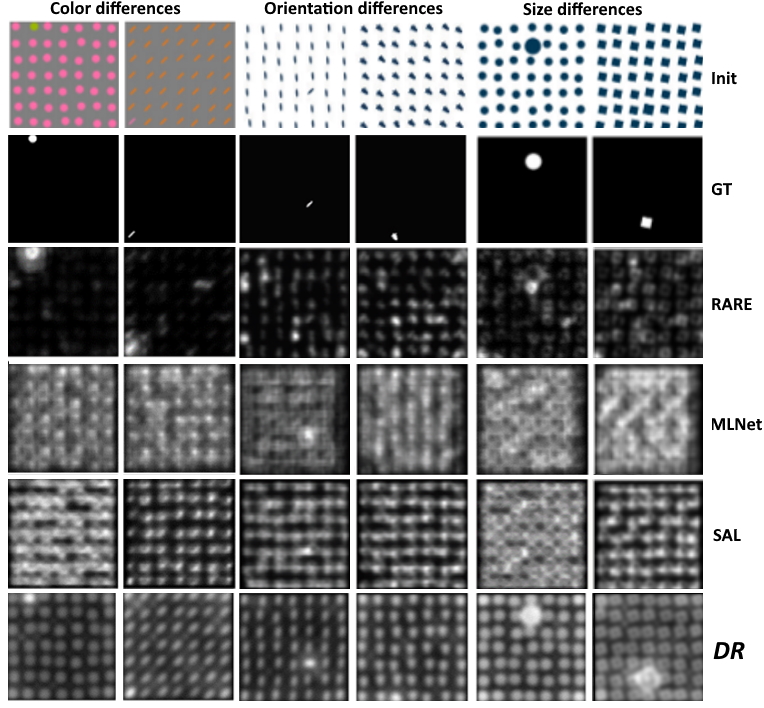}
\caption{Selected samples P\textsuperscript{3} dataset. From left to right : target difference in color, orientation, and size. From top to down : initial, ground truth, RARE2012, MLNET, SALICON, DR.}
\label{fig:fig3sup}
\end{figure}

\begin{figure}[!t]
\centering
\includegraphics[width=3.4in]{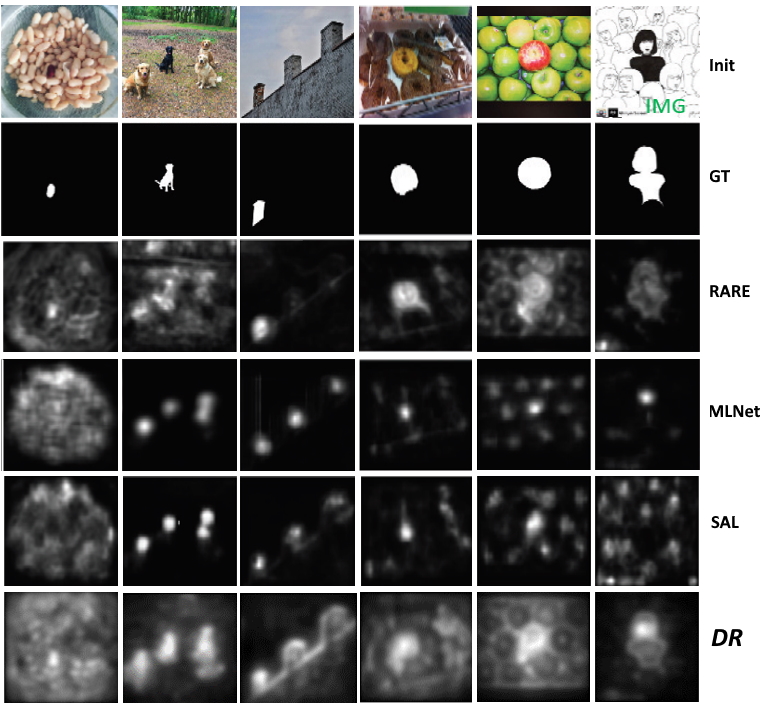}
\caption{Selected samples O\textsuperscript{3} dataset. From top to down : initial, ground truth, RARE2012, MLNET, SALICON, DR.}
\label{fig:fig4sup2}
\end{figure}

\begin{figure}[!t]
\centering
\includegraphics[width=3.4in]{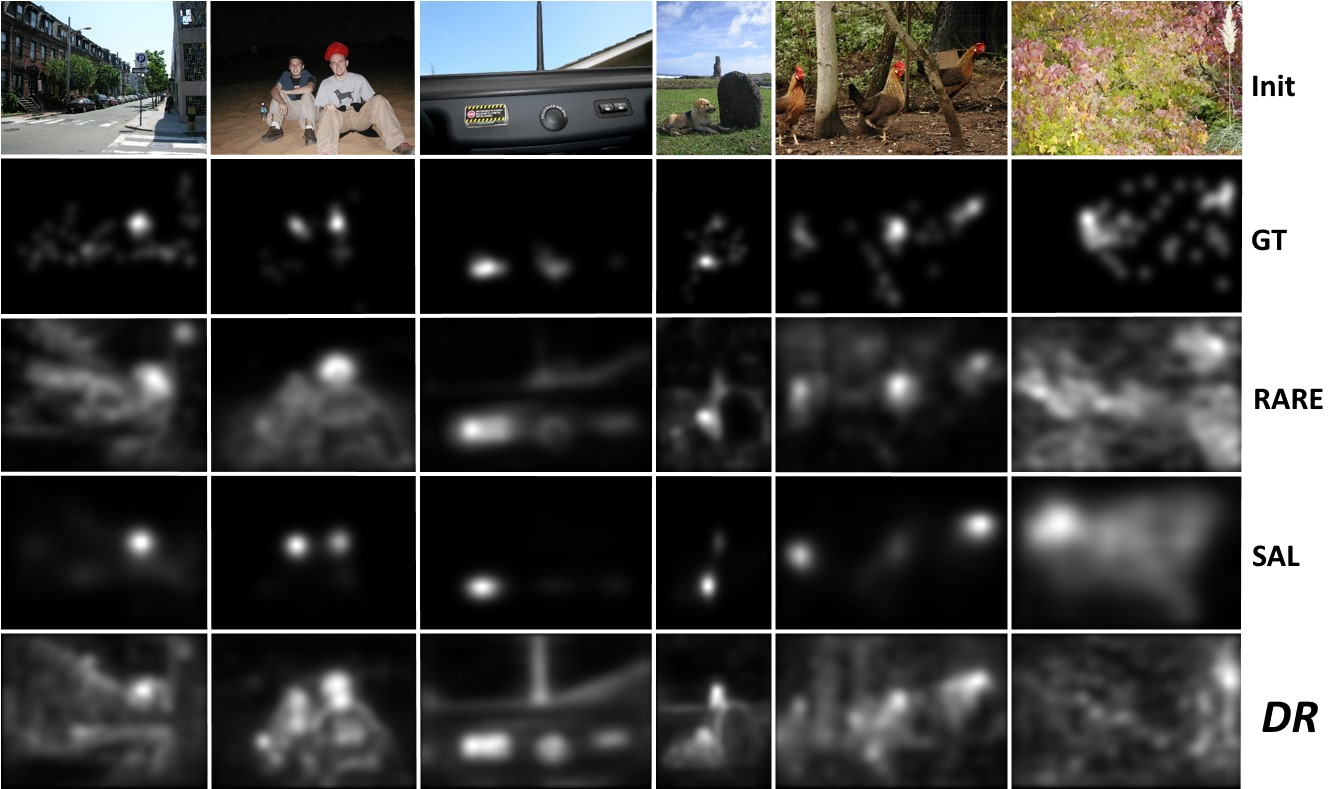}
\caption{Selected samples MIT1003 dataset. From top to down : initial, ground truth, RARE2012, SALICON, DR.}
\label{fig:figMIT}
\end{figure}

Figure \ref{fig:fig3sup} shows six samples from P\textsuperscript{3} dataset which exhibit color, orientation and size differences of the target. While distractors are still visible on DR saliency map, the targets are always correctly highlighted compared to RARE2012 which works well mainly for colors and two DNN-based models (MLNet and SALICON) which only work on one sample. Figure \ref{fig:fig4sup2} shows images from O\textsuperscript{3} dataset for different target categories (easy or difficult). Again, our model highlights the target better than the DNN-based models. DR seems equivalent in average with RARE. Figure \ref{fig:figMIT} shows images from MIT1003 dataset. DR always finds the GT focus regions (except for the right image where one GT focus is just in the middle probably due to the centred bias) but it also has details around those focus areas which might decrease its scores on MIT1003.

In overall, DeepRare2019 has the most stable behaviour performing well on both datasets while the other models might be good on some images but much less good on others. 

\subsection{Quantitative validation}
We make a quantitative validation of our model on three datasets. First on MIT1003 dataset which shows general-purpose images where learning objects is very important. This dataset is basically one which should provide advantage to DNN-based models which focus on objects instead of salient information (faces, text, etc.). 
Second, we use O\textsuperscript{3} dataset from \cite{Kotseruba2019} which also provides real life images but with odd-out-one regions. The dataset should provide similar difficulty to classical and DNN-based saliency models. 
Finally, we use P\textsuperscript{3} dataset from \cite{Kotseruba2019} which shows synthetic psycho-physical images with pop-out objects which should work better for classical saliency models. 

\subsubsection{MIT1003 dataset}
We summarize in Table \ref{tab:omit} the results of DeepRare2019 and also results coming from \cite{Kotseruba2019} for MLNet and SALICON where MLNet was trained with SALICON, P\textsuperscript{3} and O\textsuperscript{3} datasets and SALICON was trained with OSIE, P\textsuperscript{3} and O\textsuperscript{3} datasets. For other models (DeepFeat, eDN, GBVS, RARE2012, BMS, AWS), the figures come from \cite{deepFeat}. 

We remark that our model is less good than SALICON (and probably than newer models such as SAM-ResNet), but equivalent to MLNet and better than other DNN-based models. It is also better than DeepFeat and all classical models.

\begin{table}[!t]
\begin{center}
\caption{MIT1003 dataset. DFeat, eDN, GBVS, RARE2012, BMS, AWS figures come from \cite{deepFeat} and SALICON and MLNet come from \cite{Kotseruba2019}.}
\label{tab:omit}
\begin{tabular}{|c|c|c|c|c|c|c|}
\hline 
 & AUCJ & AUCB & CC & KL & NSS & SIM\tabularnewline
\hline 
SAL & 0.83 & - & \textbf{0.51} & \textbf{1.12} & \textbf{1.84} & \textbf{0.41}\tabularnewline
\hline 
\textit{DR} & \textit{\textbf{0.86}} & \textit{\textbf{0.85}} & \textit{0.48} & \textit{1.25} & \textit{1.58} & \textit{0.36}\tabularnewline
\hline 
MLNet & 0.82 & - & 0.46 & 1.36 & 1.64 & 0.35\tabularnewline
\hline
DFeat & 0.86 & 0.83 & 0.44 & 1.41 & - & -\tabularnewline
\hline 
eDN & 0.86 & 0.84 & 0.41 & 1.54 & - & -\tabularnewline
\hline
GBVS & 0.83 & 0.81 & 0.42 & 1.3 & - & -\tabularnewline
\hline 
RARE & 0.75 & 0.77 & 0.38 & 1.41 & - & -\tabularnewline
\hline 
BMS & 0.75 & 0.77 & 0.36 & 1.45 & - & -\tabularnewline
\hline 
AWS & 0.71 & 0.74 & 0.32 & 1.54 & - & -\tabularnewline
\hline 
\end{tabular}
\end{center}
\end{table}

\subsubsection{O\textsuperscript{3} dataset}
The O\textsuperscript{3} dataset uses the MSR metric defined in \cite{Kotseruba2019}. When the MSR\textsubscript{t} is higher, it is better as the target is well highlighted compared to the distractors. When MSR\textsubscript{b} is lower, it is better, it means that the maximum of the saliency of the target is higher than the one of the background. The first measure will ensure that the target is visible compared to the distractors and the second that it is visible compared to the background.  

\begin{table}[!t]
\small
\begin{center}
\caption{Comparing result between several models and DR. For MSR\textsubscript{t} higher is better, For MSR\textsubscript{b} lower is better} 
\label{tab:op3}
\begin{tabular}{|c|c|c|c|c|c|c|}
\hline 
{Model} & \multicolumn{2}{c|}{Color} & \multicolumn{2}{c|}{Non-color} & \multicolumn{2}{c|}{All targets}\tabularnewline
\cline{2-7} 
 & MSR\textsubscript{t} & MSR\textsubscript{b} & MSR\textsubscript{t} & MSR\textsubscript{b} & MSR\textsubscript{t} & MSR\textsubscript{b}\tabularnewline
\hline 
\textbf{SAM} & \textbf{1.47} & 1.46 & \textbf{1.04} & 1.84 & \textbf{1.40} & 1.52\tabularnewline
\hline 
CVS & 1.43 & 2.43 & 0.91 & 4.26 & 1.34 & 2.72\tabularnewline
\hline 
\textbf{DGII} & 1.32 & 1.55 & 0.94 & 1.95 & 1.26 & 1.62\tabularnewline
\hline 
FES & 1.34 & 2.53 & 0.81 & 5.93 & 1.26 & 3.08\tabularnewline
\hline 
\textbf{ICF} & 1.30 & 2.00 & 0.84 & 2.03 & 1.23 & 2.01\tabularnewline
\hline 
BMS & 1.29 & 0.97 & 0.87 & 1.59 & 1.22 & 1.07\tabularnewline
\hline 
\textbf{\textit{DR}} & \textit{1.14} & \textit{\textbf{0.75}} & \textit{\textbf{1.00}} & \textit{\textbf{1.00}} & \textit{1.06} & \textit{\textbf{0.89}}\tabularnewline
\hline 
\end{tabular}
\end{center}
\end{table}

Table \ref{tab:op3} shows the MSR from \cite{Kotseruba2019}  where we added DeepRare2019 at the end splitting the dataset between the images where color is a good discriminator (Color) and the others (Non-color). All models work better for targets where color is an important feature and less well for non-color. 

For MSR\textsubscript{t}(higher is better) for Color our model is less good especially compared to DNN-based models. However we can see that for Non-color images where the models fail much more DeepRare2019 has a remarkable stability being second and very close the the best one (SAM-ResNet). 

If we take into account the MSR\textsubscript{b} (lower is better), our model clearly outperforms all the others providing the best discrimination between the target and the background. DeepRare2019 is the only model with a MSR\textsubscript{b} smaller than 1 which means that in average the maximum of the target saliency is higher than the maximum of the background saliency. 

Table \ref{tab:op3o3} shows the results of the two DNN-based models also tested in Table \ref{tab:omit}. Our model outperforms both SALICON and MLNet models on both MSR\textsubscript{t} and MSR\textsubscript{b} metrics. 

\begin{table}[!t]
\begin{center}
\caption{SALICON, MLNet and DeepRare2019 rusults on O\textsuperscript{3} dataset}
\label{tab:op3o3}
\begin{tabular}{|c|c|c|}
\hline 
Model & MSR\textsubscript{t} & MSR\textsubscript{b}\tabularnewline
\hline 
\hline 
MLNet & 0.96 & 0.91\tabularnewline
\hline 
SALICON & 0.90 & 1.26\tabularnewline
\hline 
\textit{DR} & \textit{\textbf{1.06}} & \textit{\textbf{0.89}}\tabularnewline
\hline 
\end{tabular}
\end{center}
\end{table}

\subsubsection{P\textsuperscript{3} dataset}
The P\textsuperscript{3} dataset is the one which exhibits the less top-down information and it even does not have any centered bias. Naturally, for this dataset, the DNN-based models perform the worst. We will check here how DeepRare2019 deals with the data. 

First we use the average \# of fixations and found percentage metrics. Table \ref{tab:op4} shows first the results on P\textsuperscript{3} for DeepRare2019 compared with SALICON and MLNet models. Our model definitely outperforms the two DNN-based models and needs much less fixations to discover more of the targets showing here very good results.  

\begin{table}[!t]
\begin{center}
\caption{Comparing result on P\textsuperscript{3} dataset}
\label{tab:op4}
\begin{tabular}{|c|c|c|}
\hline 
Model & Avg. \# fix. & \% found \tabularnewline
\hline 
\hline 
MLNet & 42.00 & 0.44\tabularnewline
\hline 
SALICON & 49.37 & 0.65\tabularnewline
\hline 
\textit{DR} & \textit{\textbf{16.34}} & \textit{\textbf{0.87}} \tabularnewline
\hline 
\end{tabular}
\end{center}
\end{table}

\begin{figure}[!t]
\centering
\includegraphics[width=3.4in]{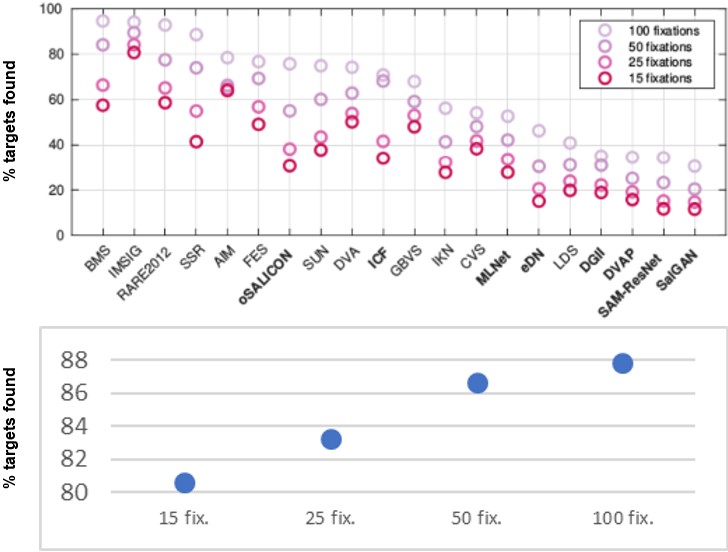}
\caption{Number of fixations vs. \% of targets detected. First row: results from \cite{Kotseruba2019} for state-of-the-art models. Second row: DR. Labels for DNN-based models are shown in bold.}
\label{fig:nb100}
\end{figure}

Figure \ref{fig:nb100} shows that compared to state-of-the-art models (top graph), our model (bottom-graph) ranges between 80 \% of targets found after 15 fixations to 88 \% target found after 100 fixations. It is possible to see that even after 15 fixations more than 80 \% of the targets are found which is much better than all tested DNN-based models and all classical models excepting IMSIG \cite{imsig2012} which has equivalent results.  

\begin{figure}[!t]
\centering
\includegraphics[width=3.4in]{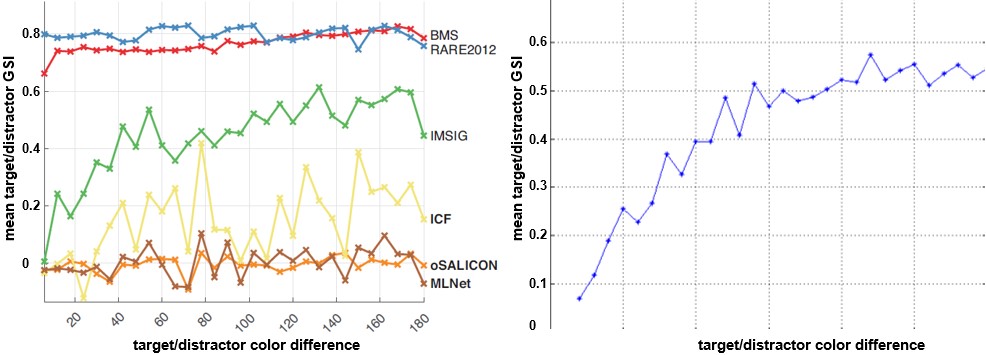}
\caption{The GSI score for color target/distractor difference. Left plot: generated by \cite{Kotseruba2019}. Right plot: DeepRare2019.}
\label{fig:gsic}
\end{figure}

\begin{figure}[!t]
\centering
\includegraphics[width=3.4in]{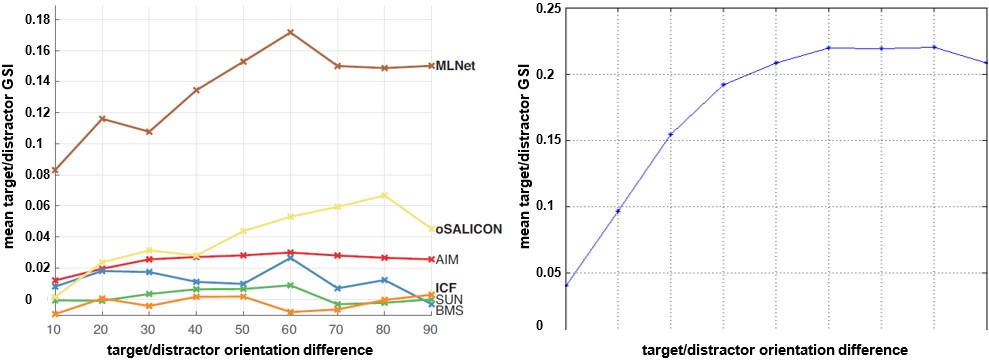}
\caption{The GSI score for orientation target/distractor difference. Left plot: generated by \cite{Kotseruba2019}. Right plot: DeepRare2019.}
\label{fig:gsio}
\end{figure}

\begin{figure}[!t]
\centering
\includegraphics[width=3.4in]{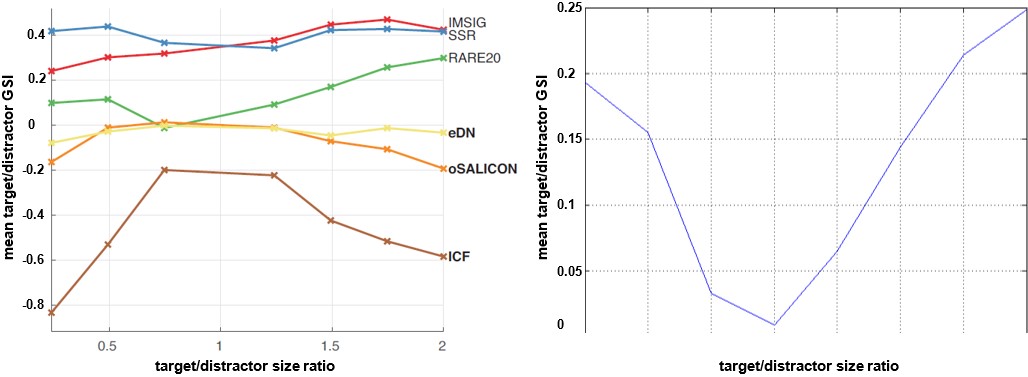}
\caption{The GSI score for size target/distractor ratio. Left plot: generated by \cite{Kotseruba2019}. Right plot: DeepRare2019.}
\label{fig:gsis}
\end{figure}

For the GSI score, figures \ref{fig:gsic}, \ref{fig:gsio} and \ref{fig:gsis} let us compare the three best classical models with the three best DNN-based models on the left and DeepRare2019 results on the right. 
For color targets (Figure \ref{fig:gsic}, right graph) we see that the maximum of GSI score for DR is 0.56 which puts our model under BMS, RARE2012 and IMSIG but much better than all the other models. 

In addition, the shape of the GSI curve exhibited by DeepRare2019 is coherent from a biological point of view: if the difference between the target color and the distractor color is small, then the model detects less well the target (left-side of the curve) than when the color of the target and background is very different (right-side of the curve). Our model is the only one to provide a biologically plausible GSI curve. 

For orientation targets (Figure \ref{fig:gsio}, right graph) we see that the maximum of GSI score is about 0.22. This makes DeepRare2019 better than any other model in terms of maximum. 

Also, the shape of the GSI curve exhibited by DeepRare2019 is again coherent from a biological point of view: if the difference between the target orientation and the distractor orientation is small (left-side of the curve), then the model detects the target less well than when target orientation is very different from the distractors (right-side of the curve). 

For size targets (Figure \ref{fig:gsis}, right graph) we see that the maximum of GSI score is about 0.25 which makes it close to RARE2012 in terms of maximum GSI. 

The shape of the GSI curve exhibited by our model is finally again coherent from a biological point of view: if the difference between the target size and the distractor size is small (center of the curve), then the model detects the target less well than when its size is very different (left and right sides of the curve). 
We can also see an asymmetry in the curve showing that it is easier for DeepRare2019 to detect target twice bigger than distractors than targets twice smaller than the distractors which is again biologically coherent.

\section{Discussion}
\label{sec:disc}
We proposed a novel saliency model called DeepRare2019 using the simplified rarity idea of \cite{rare2012} applied on the deep features extracted by a VGG16 network pretrained on ImageNet dataset. This exhibits several interesting features. 
\begin{itemize}
\item It needs no training and the default ImageNet training is enough.
\item The model is computationally efficient and is easy to run on CPU at less than one second per image.
\item Our approach is very modular, and it is very easy to adapt to any neural network architecture such as VGG19, ResNET50 or MobileNetV2 for adaptation on mobile devices such as smartphones.
\item It is possible to check each layer contribution and thus better understand the result contrary to black-box DNN-based models.
\item DeepRare2019 is very generic and stable through all kinds of different datasets where other models are sometimes better but only for one dataset and/or a specific metric but much worse for the others.
\end{itemize}

We show that this model is the most stable and generic when testing it on 3 very different datasets. It was first tested on MIT1003 where it outperforms all the classical models and most of the DNN-based models. However some DNN-based models, especially the latest ones still provide better results. We then tested DeepRare2019 on the O\textsuperscript{3} dataset, where it outperforms all the models on target/background discrimination. On target/distractor discrimination, other models perform better for Color, but our model is second on non-Color showing its stability again. Finally, on P\textsuperscript{3} dataset, our model is first ex-aequo for the target discrimination based on the number of fixations. When computing the average GSI metric our model is the only one to be in the top-three for all the features (color, orientation, size) and the only one to exhibit a GSI plot which is biologically plausible.

While one cannot expect from an unsupervised model such as DeepRare2019 to be better on MIT1003 dataset than DNN-based models which are trained and tuned on similar data, those DNN-based models are bad or even completely lost on O\textsuperscript{3} and P\textsuperscript{3} datasets. The other way around, classical models are sometimes better than DeepRare2019 on the latter datasets, but they perform much worse than DeepRare2019 on MIT1003 dataset. In addition they outperform DeepRare2019 only on specific metrics and never on all the dataset subclasses. 

\section{Conclusion}
\label{sec:ccl}
To conclude, DeepRare2019 is always the best or in the top-3 or top-4 best models in all tests we achieved. No other model is capable to be good in all datasets and their subclasses. DeepRare2019 is definitely the most stable and generic model within the tested saliency models. 

All those advantages show that deep-features-engineered models might become a good choice in visual attention field especially when the images they are applied on are special and specific eye-tracking datasets are not available or when explaining the result is of high importance.


\section*{Acknowledgments}
Supported by ARES-CCD (program AI 2014-2019) under the funding of Belgian university cooperation.



%


\printbibliography

\end{document}